%% file: main.tex
\documentclass[sigconf]{acmart}

\usepackage{caption}
\usepackage{subcaption}
\usepackage{soul}
\usepackage{enumitem}
\usepackage{xspace}

\AtBeginDocument{%
  \providecommand\BibTeX{{%
    \normalfont B\kern-0.5em{\scshape i\kern-0.25em b}\kern-0.8em\TeX}}}

\copyrightyear{2024}
\acmYear{2024}
\setcopyright{acmlicensed}\acmConference[SIGIR '24]{Proceedings of the 47th International ACM SIGIR Conference on Research and Development in Information Retrieval}{July 14--18, 2024}{Washington, DC, USA}
\acmBooktitle{Proceedings of the 47th International ACM SIGIR Conference on Research and Development in Information Retrieval (SIGIR '24), July 14--18, 2024, Washington, DC, USA}
\acmDOI{10.1145/3626772.3657948}
\acmISBN{979-8-4007-0431-4/24/07}

\begin{document}

\fancyhead{}





\title{Instruction-Guided Bullet Point Summarization of Long Financial Earnings Call Transcripts}


\author{Subhendu Khatuya}
\affiliation{%
 \institution{Indian Institute of Technology}
 \city{Kharagpur}
 \country{India}}

\author{Koushiki Sinha}
\affiliation{%
 \institution{Vellore Institute of Technology}
 \city{Chennai}
 \country{India}}

 \author{Niloy Ganguly}
\affiliation{%
 \institution{Indian Institute of Technology}
 \city{Kharagpur}
 \country{India}}

 \author{Saptarshi Ghosh}
\affiliation{%
 \institution{Indian Institute of Technology}
 \city{Kharagpur}
 \country{India}}

 \author{Pawan Goyal}
\affiliation{%
 \institution{Indian Institute of Technology}
 \city{Kharagpur}
 \country{India}}


\begin{abstract}
\input{abstract}

\end{abstract}

\if{0}


\fi



\maketitle

\input{introduction}

\input{methodology}

\input{results}

\input{conclusion}

\bibliographystyle{ACM-Reference-Format}
 \bibliography{reference}

\end{document}

%% file: abstract.tex
 While automatic summarization techniques have made significant advancements, their primary focus has been on summarizing short news articles or documents that have clear structural patterns like scientific articles or government reports. There has not been much exploration into developing efficient methods for summarizing financial documents, which often contain complex facts and figures. Here, we study the problem of bullet point summarization of long Earning Call Transcripts (ECTs) using the recently released ECTSum dataset. We leverage an unsupervised question-based extractive module followed by a parameter efficient instruction-tuned abstractive module to solve this task. Our proposed model FLAN-FinBPS achieves new state-of-the-art performances outperforming the strongest baseline with \textbf{14.88\%} average ROUGE score gain, and is capable of generating factually consistent bullet point summaries that capture the important facts discussed in the ECTs.

%% file: introduction.tex
\section{Introduction}

\begin{table}[tb]
\small
\centering
\begin{tabular}{|p{0.5\linewidth}|p{0.4\linewidth}|}
\hline
\textbf{Reference bullet-point Summary} & \textbf{Generated Questions}\\
\hline

$\bullet$ q2 non-gaap earnings per share \$0.97.

$\bullet$ sees fy revenue \$6.15 billion to \$6.21 billion.

$\bullet$ reported sales growth 16\% - 19\%.
    
&  
What is q2 non-gaap \textcolor{blue}{\textbf{earnings per share}}?

What is the \textcolor{blue}{\textbf{revenue}} of fy? 
 
What percentage of sales \textcolor{blue}{\textbf{growth}} was reported? 

\\ \hline

$\bullet$ q2 adjusted operating loss per share \$1.11.

$\bullet$ q2 same store sales rose 5.8 percent.

$\bullet$ q2 net profit 64 million usd.

& 

What is the q2 adjusted operating \textcolor{blue}{\textbf{loss}} per share? 

What is the estimate of q2 \textcolor{blue}{\textbf{sales}} rose ? 

What is the net \textcolor{blue}{\textbf{profit}} of q2? 
\\ \hline

\end{tabular}

\caption{ Sample of generated questions from reference bullet point summaries. We have highlighted the key topics such as earnings per share, revenue, loss, sales in \textcolor{blue}{\textbf{blue}.}}
\label{tab:gt_questions}
\vspace{-5mm}
\end{table}

\begin{figure*}
    \centering
\includegraphics[width=17.5cm,height=7.1cm]{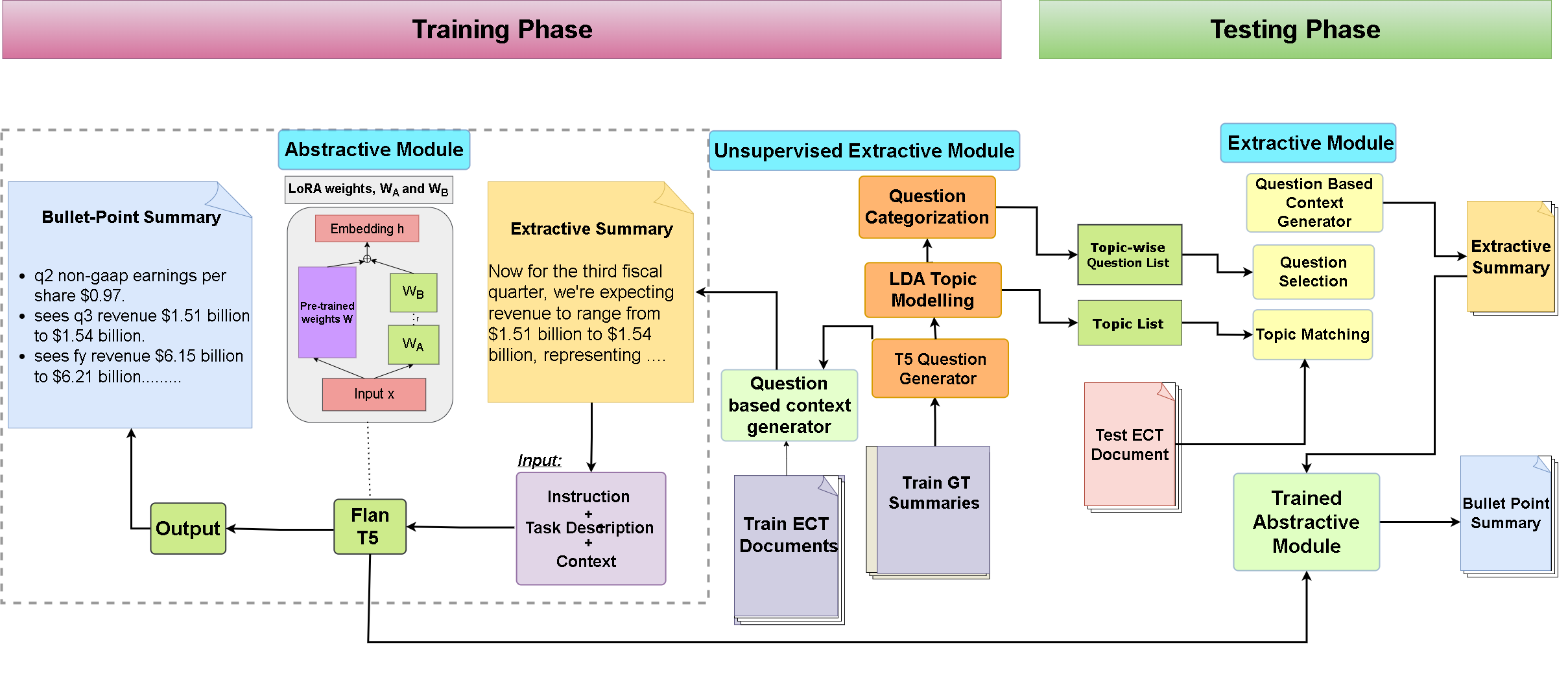}
\vspace{-6mm}
    \caption{FLAN-FinBPS Architecture - Unsupervised Extractive module takes ECT document and a list of generated questions from reference summary as input to generate the extractive context. FLAN-T5 based Abstractive module takes task specific instruction and the extractive context to generate the bullet point summary }
    \label{fig:arch}
\end{figure*}

Earnings calls are quarterly or annual conferences hosted by company executives to disclose and deliberate on the company's financial activities for the specific reporting period. 
Earnings Call Transcripts~\cite{nugent2023extractive}, commonly known as ECTs, encapsulate detailed records of the call's discussions, offering essential perspectives into the company's financial progress and future strategies. 
These transcripts serve as pivotal reference materials for investors and analysts seeking insights into the company's performance. 
ECTs are typically long unstructured documents that can be time and labour intensive to summarise by hand. Therefore development of efficient techniques for automatic ECT summarization is imperative.

Existing automatic text summarization models can perform various types of summarization, for instance: extractive summarization \cite{nallapati2017summarunner,zhong2020extractive}, abstractive summarization \cite{zhang2020pegasus,lewis2019bart}, long document summarization~\cite{zaheer2020big,beltagy2020longformer} and controllable summarization~\cite{mukherjee2020read,amplayo2021aspect}. 
However, most automatic summarization models have been primarily developed for general news articles or documents with specific layouts like government reports \cite{yang2023automatic} or scientific articles \cite{altmami2022automatic}. Financial report summarization is a fairly new area of work. Among prior works on financial document summarization, Filippova et al.~\cite{filippova-etal-2009-company} performed extractive summarization on a collection of financial news to extract information about company-specific events. 
Isonuma et al.~\cite{isonuma2017extractive} proposed a neural model for extractive summarization and applied it to financial reports. 
Leidner et al.~\cite{leidner2020summarization} presented a comprehensive survey of all existing work in financial summarization till 2020. 
More recently, the Financial Narrative Summarization shared task~\cite{el2020financial,zmandar2021financial,el-haj-etal-2022-financial} focused on extractive summarization of financial documents. 
Nugent et al.~\cite{nugent2023extractive} introduced a new unsupervised heuristic approach for extractive summarization of earning call reports.

In 2022, Mukherjee et al~\cite{mukherjee2022ectsum} proposed the ECTBPS method designed specifically for bullet point summarization  of ECT reports, and created a new benchmark dataset ECTSum\footnote{Publicly available ECTSum Dataset: \url{https://github.com/rajdeep345/ECTSum}} containing 2,425 ECT report-summary pairs. 
The target summaries in this dataset are extremely concise  bullet-point style summaries (two examples given in the first column in Table~\ref{tab:gt_questions}), facilitating 
 essential takeaways for analysts.
ECTSum~\cite{mukherjee2022ectsum} is a challenging dataset for automatic bullet point summarization primarily due to the following factors: 
(i)~it has a very high document-summary compression ratio of 103.67; 
(ii)~ECTs, with an average  document length of approx. 2.9K words, are free-form long documents as opposed to other long document summarization datasets with a fixed document layout; 
(iii)~average token count of ECTs is much more than the maximum token limit of common models such as BERT~\cite{devlin-etal-2019-bert}, T5~\cite{raffel2023exploring} or BART~\cite{lewis2019bart}.

In this paper, we propose a novel two-stage generative framework, FLAN-FinBPS, that uses a combination of unsupervised and supervised methods to produce abstractive bullet point summaries of ECT documents. 
As opposed to the previous state-of-the-art method which used supervised approaches in both stages~\cite{mukherjee2022ectsum}, we use an unsupervised question-based context generator module to produce the extractive summary in the first stage, thereby cutting down on the training time of our model. The second stage of our framework uses a supervised parameter-efficient instruction-tuned module to generate the abstractive bullet point summaries by using the extractive summary as the context. Our framework significantly improves the summarization performance over the ECTSum dataset.
Our contributions can be summarized as below:
\begin{enumerate}
    \item We propose FLAN-FinBPS \footnote{The code is available at \url{https://github.com/subhendukhatuya/FLAN-FinBPS.git}}, a novel two staged framework integrating both unsupervised and supervised methods, for more comprehensive and accurate bullet point summarization of ECT documets. The innovative unsupervised question-based extractive approach enhances the contextual understanding of the financial transcripts. 
    \item By employing an unsupervised approach in the first stage and a parameter-efficient instruction tuned Flan-T5 \cite{chung2022scaling,longpre2023flan} based generative method in the second stage, FLAN-FinBPS enhances training efficiency, making the summarization process more effective, practical and easily deployable.
    \item Our model outperforms the strongest baseline, achieving a notable \textbf{14.88\%} increase in average ROUGE score and a \textbf{16.36\%} rise in BERTScore, signifying a major enhancement in content quality. It also generates more precise numerical values, showcasing a \textbf{2.51\%} gain in Num-Prec, and produces more factually consistent summaries, demonstrating a \textbf{2.70\%} gain in SummaC\textsubscript{CONV} compared to the previous state-of-the-art method. 

\end{enumerate}

%% file: methodology.tex
\begin{table*}[tb]
    \centering
      \small
    \begin{tabular}{|l|c|c|c|c|c|c|}
        \hline
        \textbf{Model} &
        \textbf{ROUGE-1} & \textbf{ROUGE-2} & \textbf{ROUGE-L} & \textbf{BERTScore} & \textbf{Num-Prec.} & \textbf{SummaC\textsubscript{CONV}} \\
        \hline
        
        \multicolumn{7}{|l|}{\textbf{Unsupervised}} \\
        \hline
        
        LexRank \cite{erkan2004lexrank} & 0.122 & 0.023 & 0.154 & 0.638 & 1.00 & 1.00 \\
        
        DSDR \cite{he2012document} & 0.164 & 0.042 & 0.200 & 0.662 & 1.00 & 1.00 \\
        
        PacSum \cite{zheng2019sentence} & 0.167 & 0.046 & 0.205 & 0.663 & 1.00 & 1.00 \\
        \hline

        \multicolumn{7}{|l|}{\textbf{Extractive}} \\
        \hline
        
        SummaRuNNer \cite{nallapati2017summarunner} & 0.273 & 0.107 & 0.309 & 0.647 & 1.00 & 1.00 \\
        
        BertSumExt \cite{zheng2019sentence} & 0.307 & 0.118 & 0.324 & 0.667 & 1.00 & 1.00 \\
        
        MatchSum \cite{zhong2020extractive} & 0.314 & 0.126 & 0.335 & 0.679 & 1.00 & 1.00 \\
        \hline

        \multicolumn{7}{|l|}{\textbf{Abstractive}} \\
        \hline
        
        BART \cite{lewis2019bart} & 0.327 & 0.153 & 0.361 & 0.692 & 0.594 & 0.431 \\
        
        Pegasus \cite{zhang2020pegasus} & 0.334 & 0.185 & 0.375 & 0.708 & 0.783 & 0.444 \\
        
        T5 \cite{raffel2020exploring} & 0.363 & 0.209 & 0.413 & 0.728 & 0.796 & 0.508 \\
        \hline

        \multicolumn{7}{|l|}{\textbf{Long Document Summarizers}} \\
        \hline
        
        BigBird \cite{zaheer2020big} & 0.344 & 0.252 & 0.400 & 0.716 & 0.844 & 0.452 \\
        
        
        LongT5 \cite{guo2021longt5} & 0.438 & 0.267 & 0.471 & 0.732 & 0.812 & 0.516 \\
        
        LED \cite{beltagy2020longformer} & 0.450 & 0.271 & 0.498 & 0.737 & 0.679 & 0.439 \\
        \hline

        \multicolumn{7}{|l|}{\textbf{Bullet Point Summarizers}} \\
        \hline       
          ECTBPS \cite{mukherjee2022ectsum} & 0.467 & 0.307 & 0.514 & 0.764 & 0.916 & 0.518 \\
        \textbf{Flan-FinBPS} (Ours) & \textbf{0.557} & \textbf{0.376} & \textbf{0.529} & \textbf{0.889} & \textbf{0.939} & \textbf{0.532} \\

        \hline
        \midrule

        RI(\%)       & $+ 19.27\%$ &  $+ 22.47\%$ & $+ 2.91\%$ & $+ 16.36\%$ & $+ 2.51\%$ & $+ 2.70\%$\\
        \hline
    \end{tabular}
    \caption{Performance comparison of summarizers against evaluation metrics. Best scores are \textbf{bold}-ed. For \textit{Num-Prec.} and \textit{SummaC\textsubscript{CONV}}, best scores among \textit{abstractive} methods are considered (reasons in Section \ref{sec:exp:results}). Flan-FinBPS  \textbf{score the highest} on both content quality and factual consistency. The last row RI(\%) gives the relative improvement of our method (Flan-FinBPS) over the best-performing baseline (ECTBPS). 
    The baseline results are reproduced from~\cite{mukherjee2022ectsum}.}
    \vspace{-0.5em}
    \label{tab:main_res}
\end{table*}

\section{Methodology}\label{methodology}
Our proposed framework is divided into two stages: an unsupervised extractive phase, and a supervised generative  phase. The overall architecture is depicted in Figure~\ref{fig:arch}.

\subsection{Extractive Phase}
Typically, each bullet point summary highlights 3-4 crucial financial aspects of the input ECT document, such as revenue, income, earnings per share, sales, profit, equity, etc. 
This observation motivated us to initially identify the significant \textit{topics} present in a given ECT. 

We first generate a list of \textit{questions} for each ground truth summary (bullet point summary) in the train set using an unsupervised pre-trained T5 model~\cite{raffel2020exploring}. 
For each sentence $S$ in the ground truth summary, we generate a corresponding question $q$.  
Some example questions generated from the bullet point summaries are shown in Table~\ref{tab:gt_questions}.
The generation of such questions enables a more dynamic and contextually-driven approach to information extraction.

\begin{figure}[ht!]
    \centering
    \includegraphics[height=5.7cm]{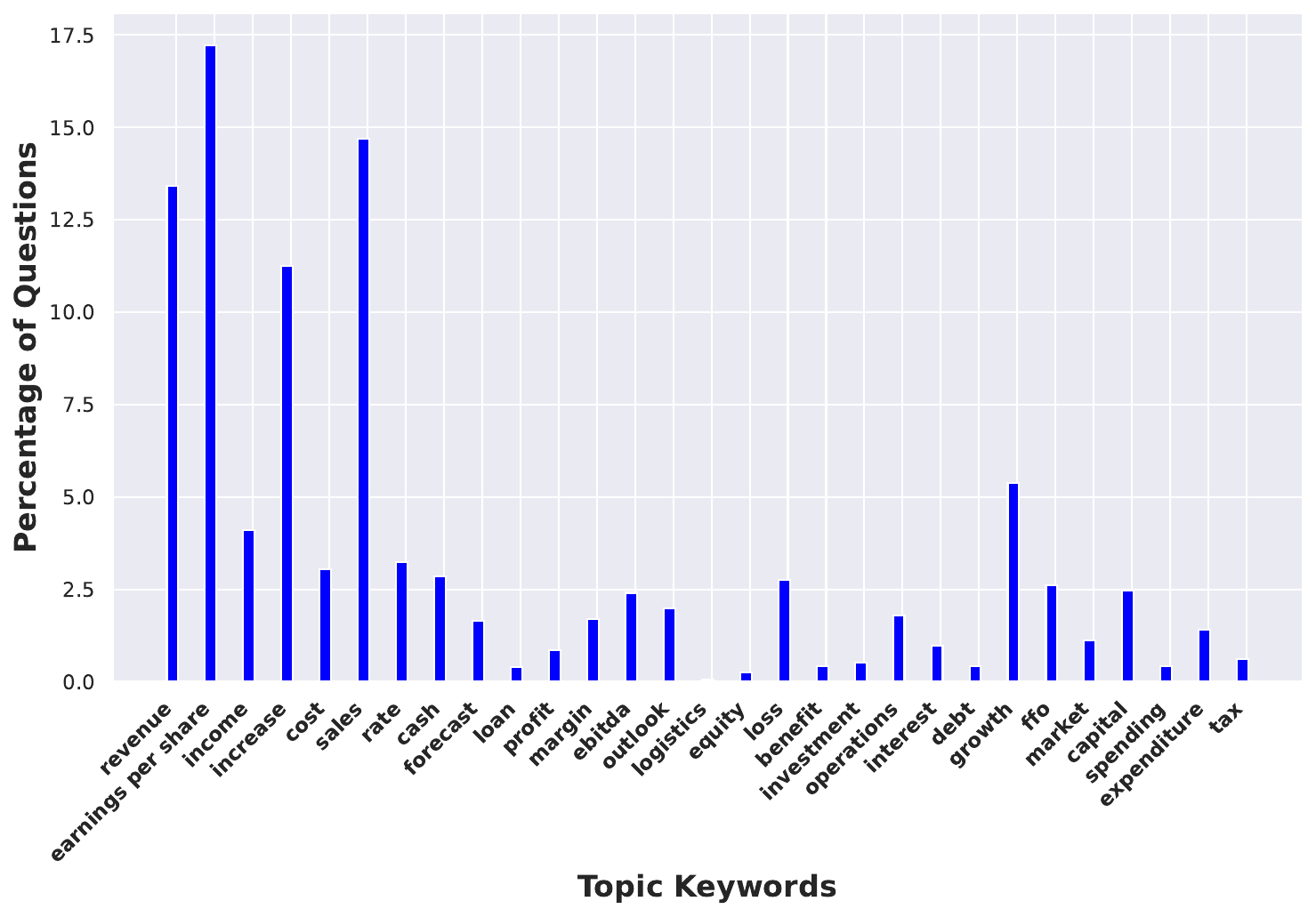}
    \caption{Distribution of questions across key topics. Revenue, earnings per Share, and sales together encompass nearly 45\% of all topics.}
    \label{fig:topic_vs_questions}
    \vspace{-4mm}
\end{figure}

Let the list of questions for each ECT $E_i$ be $Q_i$.
After removing duplicates, let the number of questions in $Q_i$ be $n$. 
We extract the top-k matched sentences based on cosine similarity score  using sentence-transformer encoder~\cite{reimers2019sentence} from each ECT for every question in its question list. 
For $E_i$,  the corresponding context will have $k*n$ sentences extracted from the ECT. This will act as input extractive context $C_i$ for the ECT $E_i$ to train the second phase of our model. We varied the value of $k$ to retrieve matched sentence from ECT and $k=3$ gave the best performance (on the validation set).
Next, we identify the key \textit{topics} from the generated questions. These topics are particularly required during the testing phase (explained in Section~\ref{sec:testing_phase}), where access to the reference summary would be unavailable. Processing all ground truth reference summaries present in the training set, we compile all the generated questions (after removing duplicates) into one master list of 3,960 questions. Then we apply LDA~\cite{jelodar2019latent} topic modelling on this list and choose the top-m ($m=30$) topics, and then categorise the questions according to the topic keywords. The value of $m$ is chosen to cover the most important financial aspects. 
Figure~\ref{fig:topic_vs_questions} represents the distribution of questions across the topics. Each topic may have multiple questions under it. 
A question is categorised under a topic if at least one keyword associated with that topic is present in it (refer Table~\ref{tab:gt_questions} where the topic-words are highlighted). Note that a question may be present under more than one category.
This approach provides a more nuanced understanding of the underlying content.

\subsection{Abstractive Phase}
In the next phase, given the extractive context $C_i$, we instruction tune FLAN-T5 \cite{chung2022scaling, longpre2023flan} with carefully-curated task-specific instructions as shown in Figure~\ref{fig:arch}, considering the bullet point summary as the target. The model is trained to condition on the modified input (i.e., extractive context with instruction) to generate the target answer one token at a time using auto-regressive decoding minimizing 
cross-entropy loss. Our choice for FLAN-T5 is based on the observation that FLAN-T5 models are pre-trained (using instruction tuning) on more than 1.8K tasks, and hence can significantly reduce the amount of fine-tuning steps required if adopted as starting checkpoints for learning new tasks. Specifically, we instruction tuned FLAN-T5-Large version with LoRA \cite{hu2021lora} technique which requires only $0.08\%$ of model parameters to be updated.

\subsection{Testing Phase} \label{sec:testing_phase}

During testing phase (refer to the right side of Figure~\ref{fig:arch}), where reference summaries cannot be utilized, we prepare the extractive context using the following approach.  

 We determine which \textit{topics} are covered by a given test ECT based on the presence of specific topic-related words. Subsequently, we employ the sentence transformer~\cite{reimers2019sentence} embeddings to identify the top-matched questions for each topic through cosine similarity, creating a list of potential questions for the given test ECT.
To construct a comprehensive context for each test ECT, we extract the top-k ($k=3$) most relevant sentences from the ECT, employing a similar approach for each identified question. This extractive summary serves as the final output of the unsupervised extractive phase, which is then passed on to the trained and instruction-tuned abstractive module for generating the final bullet-point summary.

%% file: results.tex
\section{Results and Analysis}
\label{sec:exp:results}

\noindent \textbf{Baseline methods:} We assess and compare the performance of various summarization techniques, spanning from unsupervised and extractive supervised methods to abstractive supervised models, long document summarizers, and ECTBPS~\cite{mukherjee2022ectsum} that has been designed specifically for this task. Names of the baseline methods that we considered are in Table~\ref{tab:main_res}.

\vspace{1mm}
\noindent \textbf{Dataset:} We use the ECTSum dataset~\cite{mukherjee2022ectsum} containing 2,425 ECT report-summary pairs. The dataset is split into train-val-test sets in the ratio of 7:1:2. All supervised methods are trained/fine-tuned over the train set, and all methods are evaluated over the test set.

\vspace{1mm}
\noindent \textbf{Experimental Setup:}
We instruction tune FLAN-T5-Large\footnote{\url
{https://huggingface.co/google/flan-t5-large}} model with LoRA for 10 epochs, setting \textit{rank} for the trainable decomposition as 2 with a learning rate of $5e-4$ (training time: 15 minutes/epoch, inference time: 2 minutes/sample). The input length was limited to 128, and the output to be generated was set to 60 (hyper-parameters chosen based on validation set).

\vspace{1mm}
\noindent \textbf{Evaluation Metrics:} To evaluate the content quality of model generated summaries, we consider ROUGE~\cite{lin2004rouge}, and BERTScore~\cite{zhang2019bertscore}. 
We report the F-1 scores corresponding to ROUGE-1, ROUGE-2, and ROUGE-L. Accurate reporting of monetary figures is vital in the financial domain to ensure transparency, trust, and informed decision-making among stakeholders. 
To gauge factual accuracy, we employ \textit{SummaC\textsubscript{CONV}} \cite{laban2022summac}, a newly developed NLI-based model for detecting factual inconsistencies. 
 Additionally, we report \textit{Num-Prec}~\cite{mukherjee2022ectsum} to evaluate the precision and correctness of numerical values in the summaries.

\begin{table}
\centering
\begin{tabular}{p{1.0\linewidth}  p{0.1\linewidth}}
\hline
\textbf{Reference Summary:} 
       q4 ffo per share \$0.41

       q4 earnings per share \$0.05.\\


\textbf{FLAN-FinBPS Summary:} 
\textcolor{blue} {\textbf{ffo per share \$0.41. }}
\textcolor{blue} {\textbf{q4 2020 per share \$0.05.}}
\\

\textbf{ECTBPS Summary:} q4 ffo per share \textcolor{red}{\textbf{\$0.45}}.

\textcolor{red}{\textbf{qtrly net interest expense increase by approximately \$0.03 per share}}\\
\hline
\midrule

\textbf{Reference Summary:} reports q1 sales of \$455 million.
    
    q1 adjusted earnings per share \$0.52 from continuing operations. \\

\textbf{FLAN-FinBPS Summary:} 
\textcolor{blue}{\textbf{q1 sales \$455 million.}}
\textcolor{blue}{\textbf{q1 earnings per share \$0.52 from continuing operations.}} \\

\textbf{ECTBPS Summary:} \textcolor{red}{\textbf{q3}} adjusted earnings per share \$0.52 from continuing operations. \textcolor{red}{\textbf{qtrly adjusted gross profit margin of 16.1\% versus 12.1\%.} }\\
\hline
\midrule

\textbf{Reference Summary:} 
q3 ffo per share \$0.32.

q3 adjusted ffo per share \$0.32. \\

\textbf{FLAN-FinBPS Summary:} 
\textcolor{blue}{\textbf{q3 ffo per share \underline{\$0.32}.}}
\textcolor{blue}{\textbf{q3}} \textcolor{red}{\textbf{\underline{revenue earnings}}} \textcolor{blue}{\textbf{ffo per share}} \textcolor{blue}{\textbf{\underline{\$0.32}}}. \\

\textbf{ECTBPS Summary:}
 \textcolor{blue}{\textbf{q3 ffo per share \$0.32.}}
\textcolor{red}{\textbf{qtrly rent collection rate improved to 97\% in september}}
\textcolor{red}{\textbf{sees 2021 new supply decline of 2\% to 4\% in new supply.}}
\textcolor{red}{\textbf{qtrly adjusted earnings per share \$1.00.}} \\
\hline

\midrule

\textbf{Reference Summary:} q2 adjusted earnings per share \$1.97.

q2 earnings per share \$2.19.\\

\textbf{FLAN-FinBPS Summary:} 
\textcolor{blue}{\textbf{q2 adjusted earnings per share}}\textcolor{red}{\textbf{\underline{\$2.28}}}. 

\textcolor{blue}{\textbf{q2 earnings per share \$2.19.}}  \\

\textbf{ECTBPS Summary:}
\textbf{\textcolor{red}{qtrly net sales rose 29 percent to \$348 million.}}
\textbf{\textcolor{blue}{q2 earnings per share \$2.19.}}
\textbf{\textcolor{red}{qtrly industrial sales rose 27 percent to \$231 million.}}
\\ \hline
\bottomrule

\end{tabular}

    \caption{Qualitative analysis of summaries from ECTBPS and FLAN-FinBPS. 
    Examples  show the difference between reference summary and model-generated summaries. Blue and red colour is used to indicate correct and incorrect content respectively. ECTBPS either fails to produce accurate numerical data or generates irrelevant content in most of the cases. }
    \label{tab:comparison}
    \vspace{-5mm}
\end{table}

\vspace{1mm}
\noindent {\bf Results:}
All our experimental results are detailed in Table \ref{tab:main_res}. 
 
The best scores have been highlighted in bold. Note that the Num-Prec. and SummaC\textsubscript{CONV} scores for all extractive summarizers are always 1.00 because the summary sentences are taken verbatim from the source documents. In general, the performance of the unsupervised methods when applied on their own are very poor. This signifies the domain specific nature of the ECTSum dataset and highlights the need to implement supervised methods. The methods which utilised supervised training have performed significantly better. Interestingly, Pegasus and BART, despite being initialized with weights pretrained on financial data, were outperformed by T5. 
Both T5 and long T5 models have very good factual scores.

Our model (Flan-FinBPS) demonstrates a substantial improvement over the strongest baseline ECTBPS, achieving  19.2\%, 22.4\% and 2.9\% increase in ROUGE-1, ROUGE-2 and ROUGE-L scores, respectively and an impressive 16.36\% increase in BERTScore. These advancements represent a significant leap in content quality, surpassing the current methods. 
Notably, our model excels in generating more precise numerical values, resulting in a 2.51\% increase in Num-Prec. Additionally, it delivers more factually consistent summaries, showcasing a 2.70\% gain in SummaC\textsubscript{CONV} when compared to the previous best baseline ECTBPS. 

\vspace{1mm}
\noindent {\bf Qualitative analysis of summaries:}
 A qualitative analysis of summaries generated by the strongest baseline (ECTBPS) and our proposed model is presented in Table~\ref{tab:comparison}. 
 We see that FLAN-FinBPS is mostly able to correctly capture the numerical values and the desired financial aspects (better content quality) as opposed to ECTBPS. Table~\ref{tab:comparison} presents a few instances where our model also falls short in generating the correct summary, resulting in a few hallucinated values. 
 Occasionally, the model accurately predicts financial figures but assigns them incorrectly to a different financial topic. Note that the closest baseline ECTBPS tends to exhibit a higher frequency of hallucinations than our proposed model, as also suggested by a superior Num-Prec metric for FLAN-FinBPS.

%% file: conclusion.tex
\section{Conclusion}

In this work, we propose FLAN-FinBPS for summarising earnings call transcripts into bullet point style. We showcase the effectiveness of an unsupervised, question-based extractive module, coupled with a parameter-efficient instruction-tuned generative module. This approach significantly reduces the model's training time while also attaining new state-of-the-art performance levels, surpassing the strongest baseline across all evaluation metrics. While our current application focuses on the financial domain, we aim to expand this method to other domains in our future work.